\documentclass[lettersize,journal]{IEEEtran}
\usepackage{amsmath,amsfonts}
\usepackage{algorithm}
\usepackage[noend]{algpseudocode}
\usepackage{graphicx}
\usepackage{cite}
\usepackage{tabularx}
\usepackage{multirow}
\usepackage{threeparttable}
\usepackage{booktabs}
\usepackage{makecell}
\usepackage{hyperref}
\usepackage[utf8]{inputenc}
\usepackage{pifont}
\usepackage{caption}

\begin{document}

\title{SLIDER: Sparse History-Guided Aerial Robot Target Search using Sliding Local Maps}

{
\author{Xiaolei Hou$^\dagger$, Zheng Pan$^{\dagger,*}$, Hua Lan, Zhenghao Zou, Yinhong Chen, Chenxi Zhu, \\Yang Lyu, Jinwen Hu, Chunhui Zhao%
\thanks{This work was supported in part by the Key Research and Development Program of Shaanxi Province under Grant 2024CY2-GJHX-42 and in part by the National Natural Science Foundation of China under Grants 62371398, 62293543, and 62322605.}
\thanks{$^\dagger$Equal contribution. $^*$Corresponding author: poao@mail.nwpu.edu.cn.}
\thanks{This article has been accepted for publication in IEEE Robotics and Automation Letters. Copyright 2026 IEEE. Personal use of this material is permitted.}
\thanks{Project page: \href{https://github.com/Poaos/SLIDER}{https://github.com/Poaos/SLIDER}.}
}
}

\markboth{IEEE Robotics and Automation Letters}%
{Hou \MakeLowercase{\textit{et al.}}: SLIDER: Sparse History-Guided Aerial Robot Target Search}

\IEEEaftertitletext{%
\begin{center}
\vspace{-2em}
\includegraphics[width=0.95\textwidth]{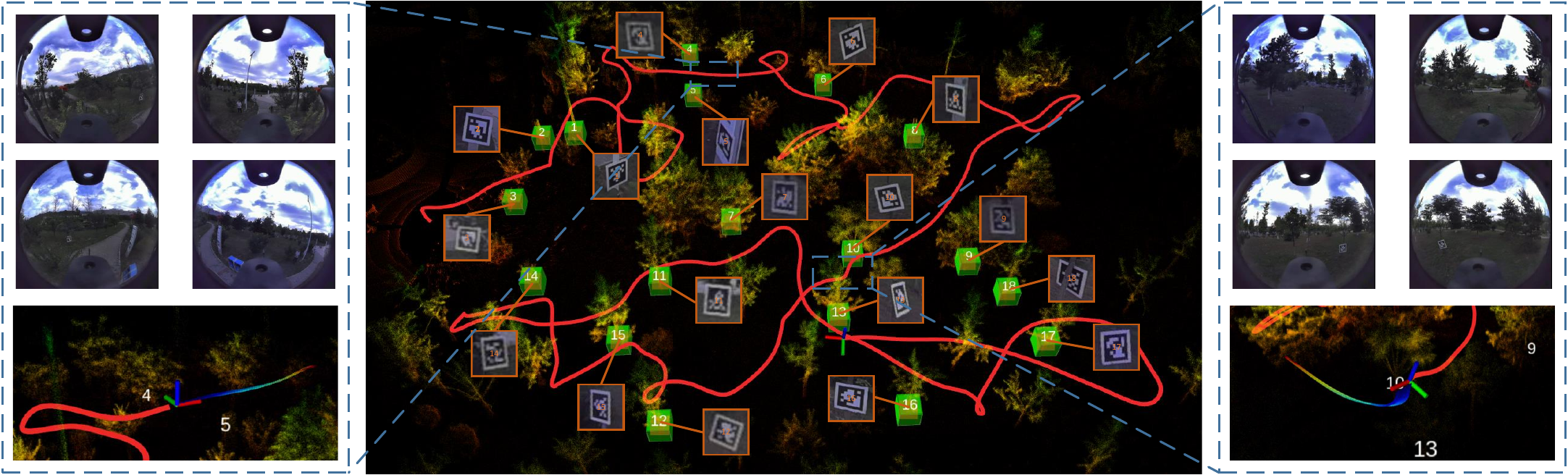}
\captionsetup{font={small}}
\captionof{figure}{Autonomous exploration of a large-scale, cluttered environment spanning several thousand square meters. The central image shows the exploration result, including the online-generated point cloud map, detected AprilTags, and the aerial robot's flight trajectory. The side images provide two snapshots during the search process, illustrating the robot's first-person view, current pose, and locally planned trajectories.}
\label{fig_cover}
\end{center}
}
\vspace{1em}

\maketitle

\begin{abstract}
Efficient exploration and target search in large-scale unknown environments remain challenging for aerial robots due to the demands of broad spatial coverage, fine-grained perception, and real-time decision-making. This paper presents SLIDER, a lightweight and memory-efficient framework that avoids reliance on globally dense maps by combining a local sliding map with sparse global history information. A novel observation quality evaluation method is proposed, leveraging historical poses and sensor models to assess point cloud data in real-time, enabling efficient frontier detection. To support scalable and responsive planning, an incremental viewpoint clustering strategy dynamically adapts to local updates, significantly reducing the number of candidate targets and decreasing computational load. A sparse global topological map is incrementally maintained to assist global planning and cost evaluation. Extensive simulations and real-world experiments demonstrate that the proposed system outperforms state-of-the-art methods in memory usage, decision latency, and search efficiency.
\end{abstract}


\section{Introduction}
\IEEEPARstart{A}{erial} robots have shown strong autonomous capabilities in tasks such as post-disaster search and rescue as well as security patrols \cite{ref-patrol}. In unknown environments, aerial robots show great potential for replacing humans in exploration and target search, which demand large-scale mapping and fine-grained perception—posing challenges in coverage, accuracy, and real-time decision-making.

To address these challenges, environmental representation serves as a critical foundation for guiding robotic perception and decision-making. Most existing exploration methods rely on occupancy grid maps \cite{ref-fuel} or octree-based structures \cite{ref-ufoexplorer} to model space, selecting candidate targets from the boundaries between known and unknown regions. Some methods further refine this strategy by sampling viewpoints near detected obstacle boundaries \cite{ref-soar}, enabling more focused observations in areas with a higher likelihood of target presence. While these strategies enhance exploration performance, grid-based methods incur increasing memory costs as the environment grows, limiting their scalability under real-time constraints \cite{ref-epic}. Although several studies have proposed more memory-efficient representations through geometric simplification \cite{ref-3m} or point cloud compression \cite{ref-gmm}, these approaches often lack sufficient spatial fidelity and adaptability for motion planning.

In addition to environmental representation, efficient target selection is crucial for exploration performance. Early methods often rely on greedy strategies that prioritize the nearest \cite{ref-frontier} or highest-gain \cite{ref-nbvp} targets, offering fast responses at the cost of global efficiency. To address this, the selection problem has been formulated as an Asymmetric Traveling Salesman Problem (ATSP), allowing for more globally optimal decisions \cite{ref-fuel}. However, solving the ATSP becomes increasingly expensive as the number of candidate targets grows, hindering real-time performance \cite{ref-eden}. Recent approaches have proposed hierarchical decision-making schemes by partitioning space \cite{ref-hphs} or clustering viewpoints \cite{ref-stars}. However, in target search tasks that require both large-scale coverage and fine-grained observation, methods based on basic region segmentation or static clustering often fail to handle the wide disparity in spatial scales and the complexity of the decision space. As a result, their computational cost remains high, making real-time performance difficult to achieve in practice.

To tackle high memory usage and computational costs in large-scale exploration, we propose SLIDER (\textbf{S}parse g\textbf{L}obal \textbf{I}nformation-\textbf{D}riven \textbf{E}fficient target sea\textbf{R}ch), a lightweight target search framework that avoids reliance on globally dense maps by maintaining a local sliding map augmented with sparse global history. The core idea is to exploit historical robot poses and the sensor model to efficiently evaluate voxel-wise observation quality, enabling rapid identification of under-explored regions without constructing a global observation map. To support real-time decision making, SLIDER employs incremental viewpoint clustering strategy to update clusters online as the environment evolves, avoiding costly global re-clustering. Additionally, a lightweight sparse topological map is incrementally constructed from selected viewpoints to guide long-horizon cost evaluation and path planning.

SLIDER is validated in simulated and real-world environments, demonstrating superior performance over state-of-the-art approaches in memory efficiency, decision latency, and search effectiveness. The main contributions of this work are:

\begin{enumerate}
 \item A lightweight framework combining a local sliding map with sparse global history, for memory- and computation-efficient target search in large-scale environments.
\item A history-aware frontier detection strategy that infers coverage quality from historical poses and sensor models, allowing rapid frontier detection without the need for a dense global observation map.
\item An efficient incremental viewpoint clustering strategy that dynamically adapts to local changes, significantly reducing candidate targets and computational costs.
\end{enumerate}
\label{intro}

\section{Related Works}

\subsection{Environment Representation for Exploration}
Environment representation serves as a core component in autonomous exploration and target search, as it encodes the robot's perception and guides its actions in unknown spaces. Traditional methods \cite{ref-fuel, ref-falcon, ref-ufoexplorer} typically rely on global map representations such as uniform grid maps or octree-based structures, where the environment is discretized into cells labeled as unknown, free, or occupied. Frontiers—free cells adjacent to unknown ones—are subsequently selected as candidate observation targets. While excelling at identifying every voxel for volumetric mapping, these methods are inefficient for object-centric tasks (e.g., target search), wasting effort on low-value open spaces. To address this mismatch, recent reconstruction \cite{ref-next, ref-soar, ref-kompis2021informed} and search \cite{ref-stars} approaches introduce the concept of surfaces—free cells adjacent to both unknown and occupied regions—to concentrate observations near obstacle boundaries. Although this improves exploration focus, such methods still rely on global occupancy grids, which impose high memory costs in large environments.

To reduce memory overhead, several methods explore more compact environmental representations. Gao et al. \cite{ref-3m} model free space using star-convex polytopes and extract triangular mesh surfaces as observation targets. While this geometric abstraction lowers memory usage, it may oversimplify complex structures, resulting in incomplete coverage. Corah et al. \cite{ref-gmm} approximate point cloud density using a Gaussian Mixture Model (GMM), achieving compression via expectation-maximization. However, GMM encodes probabilistic density rather than explicit geometry, limiting its utility for planning and requiring voxel conversion for downstream tasks. Building on this direction, Geng et al. \cite{ref-epic} proposed EPIC, a point cloud-based exploration framework that maintains obstacle surface observations and leverages an incremental kd-tree (ikd-tree) \cite{ref-ikdtree} for planning and collision checking. While EPIC avoids the heavy memory footprint of grid-based maps, it still requires maintaining both a global point cloud and a detailed observation map. As exploration scales up, the growing number of stored points increases memory usage and nearest-neighbor query costs, hindering real-time performance in large-scale environments.

Inspired by SUPER \cite{ref-super}, which maintains only a local map for fast planning, we evaluate point cloud observation quality using historical robot poses, retaining a sliding local map and sparse global history to reduce memory overhead in large-scale target search.

\subsection{Exploration Target Selection and Path Planning}
Efficiently determining visiting sequences among candidate targets is critical for exploration and target search. Early approaches, such as Yamauchi et al. \cite{ref-frontier}, adopt greedy strategies by selecting the nearest frontier. Similarly, Bircher et al. \cite{ref-nbvp} use rapidly-exploring random trees to select the node with the highest information gain. While computationally efficient, these greedy methods often result in redundant trajectories and frequent backtracking, compromising global exploration efficiency. To improve global performance, Zhou et al. \cite{ref-fuel} formulate the problem as an ATSP to minimize the cumulative cost over multiple candidate viewpoints. This reduces unnecessary revisits and has been widely adopted in later works \cite{ref-laea, ref-search25}. However, solving the ATSP becomes increasingly expensive as the number of candidates grows, limiting scalability in large-scale scenarios \cite{ref-eden}.

To alleviate this, several studies partition the space into subregions, planning a global sequence across regions followed by local planning within each \cite{ref-mtare, ref-hphs}. Zhang et al. \cite{ref-falcon} further introduce connectivity analysis within subregions, maintaining a global topological graph to guide efficient coverage planning. In target search, Luo et al. \cite{ref-stars} proposed Star-Searcher (SSearcher), a hierarchical framework that performs two-stage viewpoint clustering with history-aware decision-making, and this direction has been extended to mobile manipulators \cite{ref-heats}. However, repeated clustering at each planning step leads to high computational costs, limiting real-time performance in large environments.

To this end, we introduce an incremental viewpoint clustering strategy that dynamically updates viewpoint clusters (VCs), enabling scalable, real-time decision-making for efficient exploration and target search in large environments.

\section{Problem Statement}
We consider the problem of target search in large-scale, unknown 3D environments. Let the environment be $\mathcal{E} \subset \mathbb{R}^3$, containing a set of static targets $\mathcal{T} = \{t_1, ..., t_n\}$, where each $t_i \in \mathcal{E}$ denotes the location of a target. A robot equipped with a 3D LiDAR and a camera must autonomously explore $\mathcal{E}$ and ensure that all obstacle surfaces relevant to target detection are sufficiently observed by the onboard camera.

At time step $k$, the robot is at pose $x_k \in \mathrm{SE}(3)$ and receives a partial LiDAR scan $\mathcal{P}_k = \{p_i \in \mathbb{R}^3 \mid i = 1, ..., m\}$. The sensor's field of view (FoV) defines a local perception volume $\mathcal{V}(x_k) \subset \mathcal{E}$. The goal is to compute a motion policy $\pi = \{a_k \in \mathcal{A} \mid k = 1, ..., \tau\}$, where each $a_k$ is feasible under the system's dynamics constraints. The policy aims to explore the environment until all obstacle surfaces relevant to target detection are sufficiently observed by the onboard camera from at least one viewpoint $x_k$. Exploration terminates at the earliest step $\tau$ when full coverage is reached.

The objective is to develop a scalable, memory-efficient exploration framework that enables the robot to cover all target surfaces with minimal resource usage, without relying on a dense global map.

\section{The Proposed Approach}

\subsection{Lightweight Map Representation}
For fine-grained target search in large-scale environments, existing approaches that rely on detailed global map representations often suffer from high memory overhead and complex maintenance. Even the recent EPIC framework \cite{ref-epic} requires maintaining a dense global point cloud along with per-point observation quality, limiting scalability. To balance memory efficiency with computational performance, our system adopts a minimal and task-oriented set of map structures (Table~\ref{tab:table-map}).

\begin{table}[h]
\captionsetup{font=small} 
\caption{Map Structure Maintained \label{tab:table-map}}
\centering
\begin{tabular}{cc}
\toprule[0.8pt]
\toprule[0.8pt]
\textbf{Data}	& \textbf{Explanation}	\\
\hline
$\mathcal{M}^{\mathrm{vox}}_\mathrm{rc}$ & Local robot-centric uniform voxel grid map 
\\
$\mathcal{M}^{\mathrm{obs}}_\mathrm{rc}$ &  Local robot-centric point cloud observation quality map
\\
$\mathcal{T}_\mathrm{lidar}$ & Ikd-tree for storing local robot-centric point clouds 
\\
$\mathcal{M}_\mathrm{hist}$ & Historical robot orientation map
\\
$\mathcal{T}_\mathrm{hist}$ & Ikd-tree for storing historical robot positions 
\\
\bottomrule[0.8pt]
\bottomrule[0.8pt]
\end{tabular}
\vspace{-10pt}
\end{table}

The proposed framework leverages ROG-MAP \cite{ref-rogmap} to maintain an efficient robot-centered sliding local map \(\mathcal{M}^{\mathrm{vox}}_\mathrm{rc} = \{(v_i, s_i) \mid v_i \in \mathbb{Z}^3, s_i \in \mathcal{S}\},\) for efficient updates. Operating under an optimistic exploration paradigm, our map dispenses with the conventional three-state representation (\texttt{occupied}, \texttt{free}, \texttt{unknown}), maintaining only a compact two-state model: $\mathcal{S} = \{\texttt{occupied}, \texttt{unknown}\}$. It stores a downsampled point cloud and an inflated obstacle map, with all other voxels being implicitly unknown. By avoiding differentiation between free and unknown space, this representation reduces raycasting overhead, enabling real-time obstacle detection, local planning, and trajectory optimization.

In addition, leveraging the efficient update mechanism of $\mathcal{M}^{\mathrm{vox}}_\mathrm{rc}$, the framework also maintains a real-time local point cloud map \(\mathcal{T}_{\mathrm{lidar}} = \{p_i \in \mathbb{R}^3\}\) and a local observation quality map \(\mathcal{M}^{\mathrm{obs}}_\mathrm{rc}\ = \{(v_i,q_i) \mid v_i \in \mathbb{N}, q_i \in \mathcal{Q} \} \), where \(v_i\) is a unique voxel ID and \(q_i \in \mathcal{Q}\) encodes observation quality from LiDAR and camera. In parallel, a discretized robot orientation  map $\mathcal{M}_{\mathrm{hist}} = \{(v_i, \theta_i) \mid v_i \in \mathbb{N}, \theta_i \in \mathrm{SO}(3) \}$ and a corresponding position map $\mathcal{T}_{\mathrm{hist}}= \{p_i \in \mathbb{R}^3\}$ are maintained from the beginning of the exploration process (updated per $0.2~\mathrm{m}$ translation, matching Sec.\ref{sec-bench}, or $10^\circ$ rotation), serving the subsequent point cloud quality evaluation module. Both $\mathcal{T}_{\mathrm{lidar}}$ and $\mathcal{T}_{\mathrm{hist}}$ are implemented using ikd-tree \cite{ref-ikdtree} to support efficient nearest-neighbor and regional queries. Meanwhile, $\mathcal{M}^{\mathrm{obs}}_\mathrm{rc}$ and $\mathcal{M}_{\mathrm{hist}}$ adopt hash map structures, offering constant-time $\mathcal{O}(1)$ average complexity for insertion, lookup, and deletion.

\begin{figure}[h]
\centering
{\includegraphics[width=0.48\textwidth]{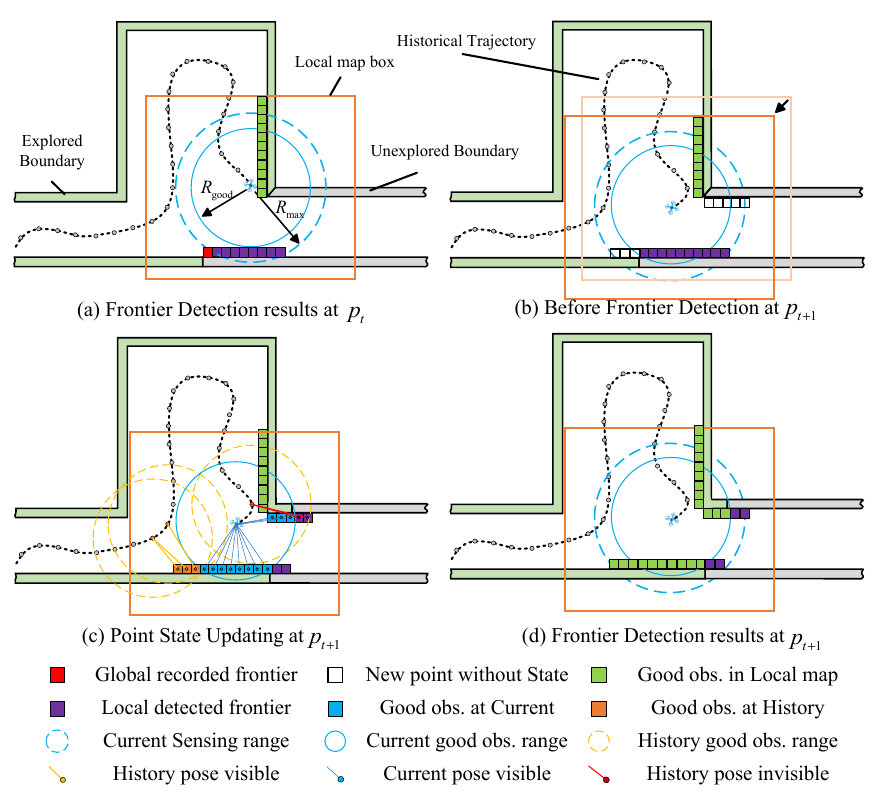}}
\captionsetup{font={small}}
\caption{{History-aware frontier detection module. The dashed blue circle indicates the maximum processed sensing range, while the solid blue and dashed yellow circles denote good observation ranges at the current and historical sensor poses. Red, white, yellow, blue, and purple voxels correspond to global frontiers, new candidates, historical good observations, current good observations, and poor observations, respectively. As the robot moves from $p_t$ to $p_{t+1}$ (a–b), voxel states are evaluated using both current and historical sensor poses (c), yielding the refined observation status in (d).}}
\label{fig_frt}
\vspace{-10pt}
\end{figure}

\subsection{History-Aware Frontier Detection and Clustering}
Traditional exploration methods maintain global maps of previously observed regions, such as occupancy grids in FALCON~\cite{ref-falcon}, voxel-wise observation distance maps in SSearcher~\cite{ref-stars}, or per-point observation metrics in EPIC~\cite{ref-epic}, which can incur significant memory and computational costs in large-scale or complex environments. To address this, we propose a lightweight frontier detection module that leverages sparse historical robot poses together with the sensor model to backward-infer observation status, enabling real-time frontier detection while reducing memory usage.

Specifically, we define a frontier as any voxel that fails to meet the sufficient-observation criteria: (i) lying within the sensor’s FoV (LiDAR or camera), (ii) being directly visible, as verified by a raycast operation, and (iii) satisfying the good observation distance threshold $r_\mathrm{lidar}$ or $r_\mathrm{camera}$. While this generic module processes real-time point clouds to detect frontiers for an individual sensor (as illustrated in Algorithm~\ref{alg-frt} and Fig.~\ref{fig_frt}), our system executes it hierarchically: voxels satisfying the LiDAR range but failing the stricter camera limits are classified as camera frontiers.

\begin{algorithm}[H]
\caption{Incremental Frontier Detection and Clustering}\label{alg-frt}
\begin{algorithmic}[1]

\Require Real-time point cloud data ${P}_{k}$, $\mathcal{M}^{\mathrm{obs}}_\mathrm{rc}$, $\mathcal{M}_\mathrm{hist}$, $\mathcal{T}_\mathrm{hist}$, global frontier cluster queue $L_\mathrm{clu}$,
\Ensure Updated ${L}_\mathrm{clu}$

\State $P_\mathrm{upd}, S_\mathrm{old} = \mathrm{ObtainUpdatedPoints}(P_{k}, \mathcal{M}^{\mathrm{obs}}_\mathrm{rc})$
\State $P_\mathrm{rem} = \mathrm{UpdateObservationbyCur}(P_\mathrm{upd}, \mathcal{M}^{\mathrm{obs}}_\mathrm{rc})$
\For {each $pt$ in $P_\mathrm{rem}$ }
	\State $p_\mathrm{nbr} = \mathrm{SearchLocalHistory}(pt, \mathcal{M}_\mathrm{hist}, \mathcal{T}_\mathrm{hist})$
	\If {$p_\mathrm{nbr} == \mathrm{null}$}
		\State $\mathrm{MarkPtFrontier}(pt, \mathcal{M}^{\mathrm{obs}}_\mathrm{rc})$
	\Else 
		\State $\mathrm{UpdateObservationbyHis}(pt, p_\mathrm{nbr}, \mathcal{M}^{\mathrm{obs}}_\mathrm{rc})$		
	\EndIf
\EndFor
\State $P_\mathrm{frt}, B_\mathrm{upd} = \mathrm{ObtainUpdateFrontier}(P_\mathrm{rem}, \mathcal{M}^{\mathrm{obs}}_\mathrm{rc}, S_\mathrm{old})$
\State $P_\mathrm{box} = \mathrm{ResetOldFrtCluster}(B_\mathrm{upd},L_\mathrm{clu})$
\State $P_\mathrm{old} = \mathrm{ReCheckObservation}(P_\mathrm{box},\mathcal{M}^{\mathrm{obs}}_\mathrm{rc},\mathcal{M}_\mathrm{hist},\mathcal{T}_\mathrm{hist})$
\State $P_\mathrm{frt} += P_\mathrm{old}, \mathrm{ClusterFrt}(P_\mathrm{frt},L_\mathrm{clu})$
\end{algorithmic}
\end{algorithm}
\vspace{-10pt}

\subsubsection{Current Observation Evaluation}
At each time step, the raw point cloud $P_k$ is filtered using the local observation quality map $\mathcal{M}^{\mathrm{obs}}_\mathrm{rc}$ to remove voxels already sufficiently observed, yielding the intermediate set $P_\mathrm{upd}$ (Line 1). The subset $S_\mathrm{old} \subset P_\mathrm{upd}$ stores voxels previously observed but not yet sufficient, shown as purple voxels in Fig.~\ref{fig_frt}(a--b). The observation quality of each voxel in $P_\mathrm{upd}$ is evaluated based on the current robot pose. Voxels failing any criterion are added to the remainder set $P_\mathrm{rem}$ for subsequent processing (Line~2), while the others are marked as sufficiently observed and included in $\mathcal{M}^{\mathrm{obs}}_\mathrm{rc}$ (blue voxels in Fig.~\ref{fig_frt}(c)). Since $P_k$ comes directly from sensor measurements, explicit ray-casting for occlusion checking is unnecessary, simplifying computation.

\subsubsection{Historical Pose-Based Observation Inference}
Following the initial update, voxels in the remainder set $P_\mathrm{rem}$ are further evaluated (Lines 3--8). For each voxel, the algorithm searches for any historical pose $p_\mathrm{nbr} = \{p_i, \theta_i\}$ within the predefined good-observation range and sensor FoV, where $p_i \in \mathcal{T}_\mathrm{hist}$ is used as a key to query the associated orientation $\theta_i \in \mathcal{M}_\mathrm{hist}$. If at least one such pose satisfies the visibility condition (no occlusion), the voxel is labeled as sufficiently observed and added to $\mathcal{M}^{\mathrm{obs}}_\mathrm{rc}$, shown as yellow voxels in Fig.~\ref{fig_frt}(c). By avoiding dense global maps, this backward-inference strategy reduces memory overhead and supports real-time performance. Voxels that remain insufficiently observed are labeled as frontier and also added to $\mathcal{M}^{\mathrm{obs}}_\mathrm{rc}$.

\subsubsection{Frontier Extraction and Clustering}  \label{sec-ftrcluster}
From the remaining insufficiently observed voxels, newly detected frontier voxels $P_\mathrm{frt}$, which are still labeled as frontier and not yet part of the existing frontier set, are extracted. An axis-aligned bounding box $B_\mathrm{upd}$ is then constructed to cover the newly detected frontiers, defining the update region (Line 9). The queue $L_\mathrm{clu}$, which maintains all clusters and their bounding boxes, is then updated; clusters whose bounding boxes intersect $B_\mathrm{upd}$ are reset. Voxels within these reset clusters that remain in the frontier state are collected and denoted as $P_\mathrm{box}$ (Line 10). To account for sensor noise and limited FoV, the observation quality of voxels in $P_\mathrm{box}$ near the robot's current position is re-evaluated, producing a refined subset $P_\mathrm{old}$ (Line 11). Finally, the aggregated frontier voxels $P_\mathrm{frt} \cup P_\mathrm{old}$ are clustered to update the global frontier cluster (FC) queue $L_\mathrm{clu}$ (Line 12).

However, distance-based FC may group frontier voxels with inconsistent surface orientations. We therefore use a normal-aware metric $d_\mathrm{sim}(f_i,f_j)=\lambda_1|p_i-p_j|_2-\lambda_2 n_i^\top n_j$, where $p_i$ and $n_i$ denote the position and surface normal of $f_i$, respectively, and $\lambda_1,\lambda_2>0$ are weighting coefficients.

\begin{figure*}[!t]
\centering
{\includegraphics[width=0.85\textwidth]{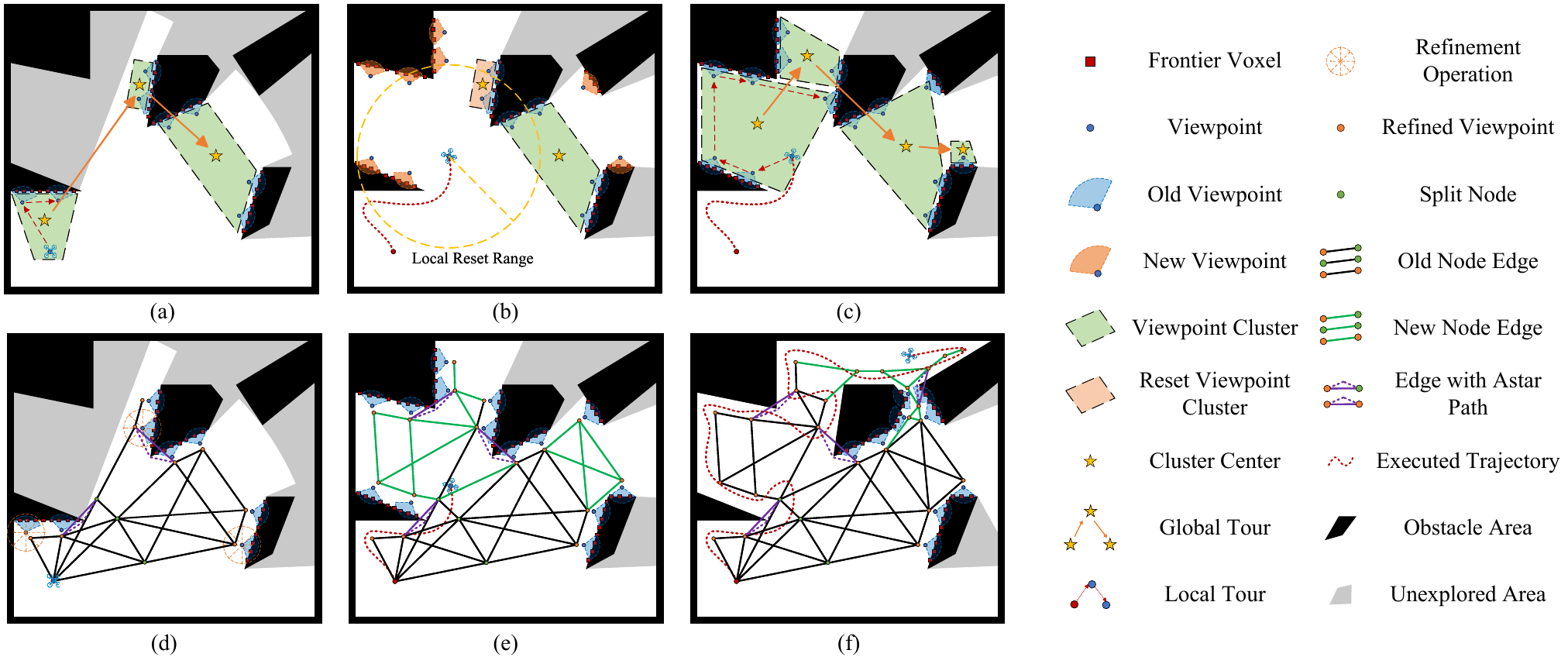}}
\captionsetup{font={small}}
\caption{{Incremental viewpoint clustering and sparse topological map construction. As the robot moves, new FCs are generated, while invalid ones (containing invalid FC IDs) and those centered near the current pose are reset (light orange box, (a–c)). The global and local visiting orders are updated accordingly (yellow and red arrows sequences). Newly generated viewpoints are refined toward traversable regions (d), and edges are maintained by splitting overly long ones with split nodes and adding low-cost connections between nearby nodes (d–f).}}
\label{fig_vp_topo}
\vspace{-10pt}
\end{figure*}
\subsection{Incremental Viewpoint Clustering for Target Generation} 
In target-oriented exploration, frontiers include not only free-space boundaries but also insufficiently observed obstacle surfaces, which expand candidate viewpoints and complicate real-time decision-making. Notably, SSearcher \cite{ref-stars} reclusters all FCs at every planning step, requiring heavy ray-casting as the environment and viewpoint set grow. This results in high computational costs and limits responsive decision-making, as later shown in Section~\ref{sec-bench}. To address this limitation, our method updates only newly added or modified FCs, avoiding redundant computation, and enabling scalable real-time target generation.

\subsubsection{Viewpoint Generation}
Effective surface exploration relies on generating high-quality sensor viewpoints. Traditional uniform sampling often ignores surface geometry, yielding suboptimal coverage. We instead exploit the consistent surface normals within FCs to sample viewpoints aligned with obstacle surfaces. Each viewpoint is first generated by applying a template transformation at the cluster centroid, oriented along its dominant normal, and then prioritized according to the number of observable frontiers.

\subsubsection{Incremental Viewpoint Clustering}
In Algorithm~\ref{alg-frt}, we track FCs that are either reset or newly added, and maintain the global set of viewpoint clusters (VCs) incrementally. Each VC consists of a group of mutually visible viewpoints and the corresponding FC IDs. A new viewpoint is merged into an existing VC if it lies within a distance threshold $r_\mathrm{vc}$ from the cluster center and satisfies visibility constraints. The update proceeds as follows: first, VCs containing invalid FC IDs, and those whose centers lie within a radius $r_\mathrm{nbr}$ of the robot’s current position $p_\mathrm{c}$, are reset, as shown in Fig. \ref{fig_vp_topo} (b). Next, $p_\mathrm{c}$ is inserted as a virtual viewpoint to initialize the first new VC, ensuring that subsequent target generation remains naturally connected to the current pose, thereby avoiding trajectory discontinuities and improving local path smoothness. Finally, existing VCs are incrementally extended, remaining viewpoints are clustered into new VCs, and the inter-cluster traversal costs are updated, yielding a dynamic structure that reduces the decision space and enables faster planning and real-time exploration in complex environments.
\subsection{Sparse Topological Map Construction} 
Prior approaches typically construct sparse topological maps by uniformly sampling nodes in free space or extracting nodes from a Voronoi-based skeleton graph, but the former often produces redundant nodes and the latter can be computationally expensive. Moreover, such methods generally connect only fully visible nodes, which may break connectivity in complex environments. To address these issues, we use informative viewpoints from updated FCs as candidate nodes, naturally covering structurally important regions while reducing redundancy and computational overhead. Edges between nodes are added whenever a low-cost optimistically feasible path exists, ensuring a sparse yet well-connected topological representation in cluttered or intricate spaces.

\subsubsection{Candidate Node Refinement}
For each selected viewpoint $vp_i$, a local refinement is performed using the sliding map, shifting the node toward open and traversable regions while retaining coverage near obstacles. Rays are cast along directions sampled via the Fibonacci sphere, terminating at obstacles or a predefined range limit, and the mean position of all ray endpoints defines a refined pose $vp'_i$, as a refined viewpoint in Fig.~\ref{fig_vp_topo}(d).

\subsubsection{Edge Formation and Topology Maintenance}
We construct a sparse topological map $G = \{V, E\}$, where $V$ denotes the refined viewpoint positions and $E$ represents the interconnecting edges. To balance sparsity and coverage, especially in complex environments where straight-line connections are often infeasible, edges are initially added between two nodes if an optimistically feasible path exists and its cost does not exceed $k_\text{rate}$ times their Euclidean distance. These edges are dynamically pruned if obstructed by new observations, ensuring the topological map provides reliable global guidance.

\subsection{Trajectory Planning with Global Guidance}
To balance global exploration efficiency and local responsiveness, we adopt a hierarchical planning strategy. At the global level, the algorithm determines a visiting sequence among VCs and refines the order of FCs within the selected VC, thereby narrowing the decision space and yielding a concrete navigation target. At the local level, trajectories toward this target are generated by the SUPER planner \cite{ref-super}, which efficiently produces dynamically feasible paths directly from point cloud data under cluttered constraints. 

Constrained by its local sliding map, SUPER is susceptible to local minima (e.g., U-shaped traps) during long-range navigation. To resolve this, our framework extracts a global path from the sparse topological map and truncates it to the local planning horizon. This immediate segment is then fed to SUPER as a local reference path, guiding the UAV around large-scale obstacles that exceed its local perception.

\begin{figure*}[!t]
\centering
{\includegraphics[width=0.93\textwidth]{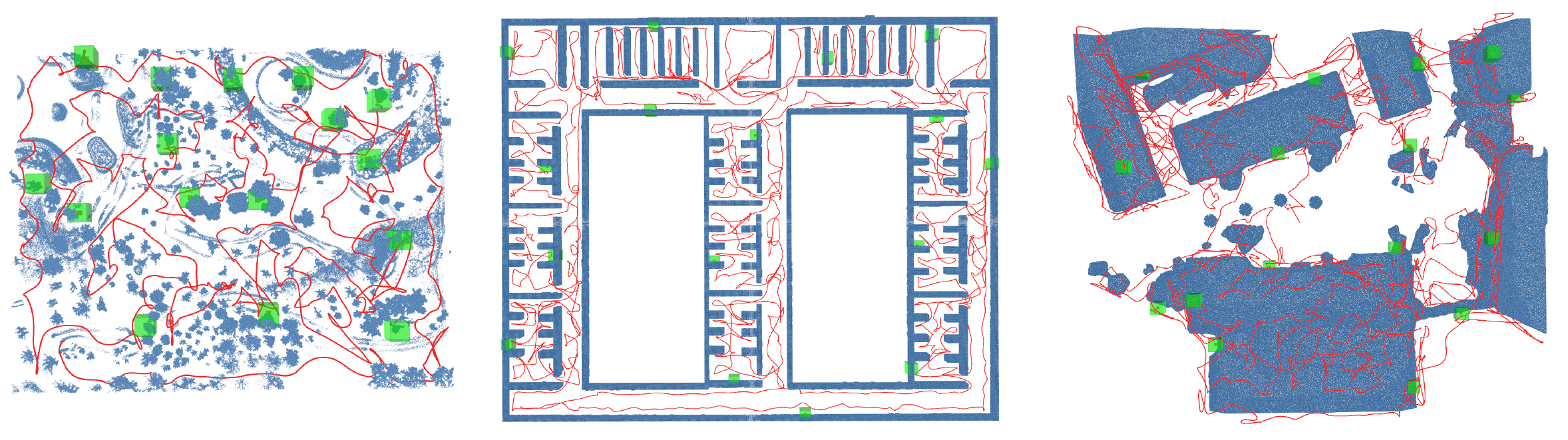}}
\captionsetup{font={small}}
\caption{The trajectories of the proposed method in forest, garage and campus scenes. The red line represents the trajectory of the aerial vehicle, and the green boxes indicate the detected targets.}
\label{fig_exp_sim1}
\vspace{-10pt}
\end{figure*} 

\section{EXPERIMENTS} \label{sec-exp}

\subsection{Simulation Setup}  \label{sec-sim}
 To evaluate the proposed method, we constructed three large-scale 3D environments in the MARSIM \cite{ref-marsim}: a forest ($90 \times 70 \times 4~\mathrm{m}^3$), a garage ($192 \times 156 \times 4~\mathrm{m}^3$), and a campus ($140 \times 120 \times 27~\mathrm{m}^3$). In each environment, 16 target objects to be searched were randomly distributed near obstacle surfaces. 

The simulated aerial robot was equipped with a MID360 LiDAR ($360^\circ$ horizontal, $[-7^\circ, 52^\circ]$ vertical FoV). As MARSIM lacks RGB modeling, a virtual sensor was used to approximate the real-world perception setup, configured based on the Seeker Omni camera  ($360^\circ$ horizontal, $[-50^\circ, 50^\circ]$ vertical FoV). These sensors provide geometric information used for visibility checking and observation quality updates. To approximate perception without image data, we adopted a geometry-based observation model in which targets were defined by 3D position and normal. A target was considered detected if it fell within the camera’s FoV and the reliable range $r_\mathrm{good}$.

\begin{figure*}[!t]
\centering
{\includegraphics[width=0.92\textwidth]{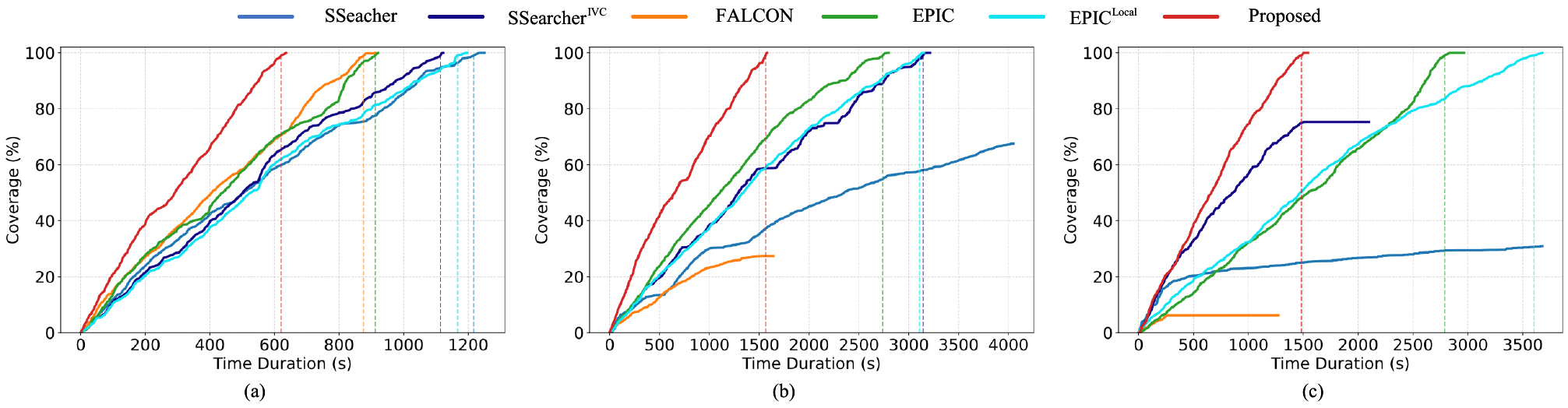}}
\captionsetup{font={small}}
\caption{The exploration progress of all the three state-of-the-art benchmarks and the proposed method in (a) forest, (b) garage and (c) campus scenes.}
\label{fig_exp_cov}
\vspace{-10pt}
\end{figure*}

\vspace{-5pt}
\subsection{Benchmark Comparisons}  \label{sec-bench}
We compare SLIDER against three representative, publicly available baselines: SSearcher~\cite{ref-stars}, FALCON~\cite{ref-falcon}, and EPIC~\cite{ref-epic}. To ensure a fair evaluation tailored to target search, observation models were unified with ours while preserving core strategies. Specifically, SSearcher’s visibility checks were aligned with our FoV constraints; FALCON was adapted to use LiDAR input with a map update range of $r_\mathrm{good}$ and spatial partition size of $30~\mathrm{m}$ for stability; and EPIC’s angular constraint was replaced with camera visibility.

Across all algorithms, the maximum LiDAR update range was limited to $13.0~\mathrm{m}$, with an observation quality resolution of $0.2~\mathrm{m}$ for both grid maps and point clouds. The reliable perception range was set to $r_\mathrm{good} = r_\mathrm{camera} = 5.0~\mathrm{m}$ and $r_\mathrm{lidar} = 12.0~\mathrm{m}$. The parameters for VCs were set to $r_\mathrm{vc} = r_\mathrm{nbr} = 10.0~\mathrm{m}$. Robot dynamics were constrained by $v_\mathrm{max} = 5.0~\mathrm{m/s}$ and $a_\mathrm{max} = 5.0~\mathrm{m/s^2}$. For SLIDER, a ROG-MAP of $0.1~\mathrm{m}$ resolution was maintained, with map sizes of $30~\mathrm{m} \times 30~\mathrm{m} \times 8~\mathrm{m}$ for the {Forest} and {Garage} scenarios, and $30~\mathrm{m} \times 30~\mathrm{m} \times 30~\mathrm{m}$ for the {Campus} scenario. All simulations were run on a laptop with an Intel i5-12500H CPU and $32~\mathrm{GB}$ RAM. Each algorithm was executed five times per scenario, with results summarized in Table~\ref{tab-cmp} and Fig.~\ref{fig_exp_cov}.

The experimental results reveal clear performance disparities among the evaluated algorithms, particularly in large-scale scenarios. In the small, uniformly structured Forest environment, all algorithms successfully completed exploration and identified all targets. FALCON’s connectivity-based coverage strategy yields slight efficiency gains in small environments but performs poorly in larger scenarios such as Garage and Campus, where the lack of directional guidance and the overhead of maintaining a connectivity topology increase computational costs, ultimately degrading performance and causing task failures. In contrast, SSearcher achieves better coverage due to its obstacle-focused method, yet repeated viewpoint clustering incurs heavy computational costs that limit real-time responsiveness. EPIC performs reliably across scenarios by operating directly on point cloud data; however, as the global ikd-tree grows (with nearest neighbor search complexity $O(\log n)$), the computational cost of candidate targets increases significantly, undermining decision efficiency.

Consequently, SLIDER leverages a lightweight map representation and incremental viewpoint clustering to reduce computational burden and enable real-time decision making. Quantitative results show a 30--50\% reduction in total exploration time, a 29--60\% increase in average speed, and minimal computational cost, while all targets were identified.

\vspace{-5pt}
\subsection{Ablation Experiments}
\subsubsection{Memory Consumption}

We benchmark memory efficiency against EPIC, a lightweight baseline significantly more compact than standard grid-based methods (e.g., SSearcher, FALCON). To isolate the impact of global versus local map structures, our variant $\mathrm{EPIC^{Local}}$ restricts only EPIC's global ikd-tree to match our sliding local map scheme, strictly retaining its global observation information.

As shown in Table~\ref{table-memory}, localizing EPIC’s map representation substantially reduces its memory usage (e.g., from 293.2~MB to 24.0~MB in the Campus scenario). While $\mathrm{EPIC^{Local}}$ achieves this reduction, it also incurs a noticeable degradation in exploration efficiency (Table~\ref{tab-cmp}), suggesting that EPIC’s planning is closely tied to globally consistent map representations. Even under this favorable setting, the proposed method consistently maintains a more compact environment representation, achieving 4--7$\times$ lower Env. Total memory than both EPIC and $\mathrm{EPIC^{Local}}$ across all scenarios. Although an additional ROG-MAP is maintained for motion planning, its memory overhead remains modest.

\begin{table*}[t]
\centering
\captionsetup{font=small}
\caption{Results of Benchmark Comparisons}
\label{tab-cmp}
\renewcommand\arraystretch{1.5}
\resizebox{\textwidth}{!}{%
\begin{threeparttable}
\begin{tabular}{c c c c c c c c c c}
\toprule[0.8pt]
\toprule[0.8pt]
\textbf{Scene} & \textbf{Method} & \textbf{\makecell{Exp. Tm. (s)}} & \textbf{\makecell{Flt. Dist. (m)}} & \textbf{\makecell{Avg. Vel. (m/s) }}& \textbf{\makecell{Topo. Tm. (s)}}& \textbf{\makecell{Can. Tm. (s)}} & \textbf{\makecell{Sel. Tm. (s)}} & \textbf{\makecell{Compl. (\%)}} \\ 
\hline
\multirow{6}{*}{Forest}
                        & SSearcher \cite{ref-stars}   & 1250.15 & 2352.45  & 1.88  & - & 0.518  & 0.128 & 100 \\ 
                        & $\mathrm{SSearcher^{IVC}}$  & 1122.92 & 2097.76  & 1.87  & - & 0.117  & 0.103 & 100 \\ 
                        & FALCON \cite{ref-falcon}  & 912.51 & 1811.34 & 1.99  & 0.109  & 0.039 & 0.124 & 100 \\ 
                        & EPIC  \cite{ref-epic}  & 921.68 & 1889.34 & 2.05 & 0.025 & 0.103 & 0.017 & 100  \\ 
                        & $\mathrm{EPIC^{Local}}$  & 1196.58 & 2461.63 & 2.06 & 0.028 & 0.140 & 0.017 & 100  \\ 
                        & Proposed & \textbf{636.15} & \textbf{1691.74} & \textbf{2.66} & \textbf{0.019} & \textbf{0.030} & \textbf{0.014} & 100  \\ 
\hline
\multirow{6}{*}{Garage}
                        & SSearcher  \cite{ref-stars}   & - & -  & 0.61  & -  & 3.193 & 0.201  & 63 \ding{55} \\ 
                        & $\mathrm{SSearcher^{IVC}}$   & 3212.42 & 4977.36  & 1.55  & - & 0.097  & 0.032 & 100 \\ 
                        & FALCON  \cite{ref-falcon}  & - & - & 1.01 & 0.504 & 0.095 & 0.913 & 25 \ding{55}
\\ 
                        & EPIC \cite{ref-epic} & 2798.49 & 5533.74 & 1.98 & 0.023 & 0.085 & 0.035 & 100  \\ 
                        & $\mathrm{EPIC^{Local}}$   & 3150.53 & 6135.95 & 1.95 & 0.025 & 0.160 & 0.039 & 100  \\ 
                        & Proposed & \textbf{1577.60} & \textbf{4893.98} & \textbf{3.12} & \textbf{0.009} & \textbf{0.022} & \textbf{0.010} & 100  \\ 
\hline
\multirow{6}{*}{Campus}
                        & SSearcher \cite{ref-stars}    & - & - & 0.21 & - & 9.134 & 2.655 & 31 \ding{55} \\ 
                        & $\mathrm{SSearcher^{IVC}}$   & - & - & 1.08 & - & 0.462 & 0.535 & 75 \ding{55} \\
                        & FALCON  \cite{ref-falcon}  & - & - & 0.56 & 0.991 & 0.055 & 0.256 & 19 \ding{55} \\ 
                        & EPIC  \cite{ref-epic} & 3213.95 & 5462.52 & 1.81 & 0.036 & 0.273 & 0.051  & 100  \\ 
                        & $\mathrm{EPIC^{Local}}$   & 3801.62 & 6977.65 & 1.84 & 0.041 & 0.391 & 0.051 & 100  \\ 
                        & Proposed & \textbf{1611.84} & \textbf{4689.24} & \textbf{2.91} & \textbf{0.010} & \textbf{0.034} & \textbf{0.009} & 100  \\ 

\bottomrule[0.8pt]
\bottomrule[0.8pt]
\end{tabular}

\begin{tablenotes}
{\item \footnotesize \textbf{Note:}
The table summarizes key performance metrics, including exploration time (Exp. Tm.), flight distance (Flt. Dist.), average velocity (Avg. Vel.), average time for topological map maintenance (Topo. Tm.), average time for candidate target generation (Can. Tm.; including viewpoint generation and clustering), average time for target selection (Sel. Tm.), and search completeness (Compl.; measured by the number of recognized targets). }
\end{tablenotes}

\end{threeparttable}
}
\vspace{-0.2cm}
\end{table*}

\begin{table}[!t]
\centering
\captionsetup{font=small}
\caption{Memory Consumption Comparisons}
\label{table-memory}
\renewcommand\arraystretch{1.1}
\setlength{\tabcolsep}{0pt} 

\begin{threeparttable}
\footnotesize
\begin{tabular*}{\linewidth}{@{\extracolsep{\fill}} c c c c c c c c c}
\toprule[0.8pt]
\bottomrule[0.8pt]

\textbf{Scene} & \textbf{Method} & \textbf{\makecell{Obs. \\ Map}} & \textbf{\makecell{Hist. \\ Tree}} & \textbf{\makecell{Ori. \\ map}} & \textbf{\makecell{Env. \\ Total}} & \textbf{\makecell{Map \\ Tree}} & \textbf{\makecell{Rog \\ Map}} & \textbf{\makecell{Mem. \\ Total}}\\ 
\hline
\multirow{3}{*}{Forest}
    & EPIC \cite{ref-epic} & 7.93 & - & - & 7.93      & 193.4 & - & 193.4 \\
    & $\mathrm{EPIC^{Local}}$ & 8.24 & - & - & 8.24   & 6.8 & - & \textbf{15.0} \\
    & Proposed & \textbf{0.36} & 0.48 & 0.56 & \textbf{1.40}  & \textbf{5.6} & 16.6 & {23.6} \\
\hline
\multirow{3}{*}{Garage}
    & EPIC \cite{ref-epic} & 22.38 & - & - & 22.38  & 272.7 & - & 272.7\\
    & $\mathrm{EPIC^{Local}}$ & 22.39 & - & - & 22.39      & 5.2 & - & 27.6 \\
    & Proposed & \textbf{0.31} & 1.20 & 1.40 & \textbf{2.92}   & \textbf{3.9} & 16.6 & \textbf{23.5} \\
\hline
\multirow{3}{*}{Campus}
    & EPIC \cite{ref-epic} & 14.22 & - & - & 14.22     & 293.2 & - & 293.2\\
    & $\mathrm{EPIC^{Local}}$ & 16.24 & - & - & 16.24      & 7.8 & - & \textbf{24.0} \\
    & Proposed  & \textbf{0.30} & {1.44} & {1.68} & \textbf{3.42}  & \textbf{4.2} & 43.1 & {50.7}\\
\toprule[0.8pt]
\bottomrule[0.8pt]
\end{tabular*}

\begin{tablenotes}
\item \footnotesize \textbf{Note:} Columns from left to right indicate the memory usage in MB (megabytes): observation quality map, historical position ikd-tree, historical orientation map, environment representation (sum of the previous three), environment point cloud ikd-tree, robot-centric voxel grid map (ROG-MAP) and the total memory usage of all modules.
\end{tablenotes} 
\end{threeparttable} 
\vspace{-0.2cm}
\end{table}

\subsubsection{Frontier Detection}
We compare the proposed local-only frontier detection method ($\mathrm{F}^\mathrm{local}$) with a baseline that maintains a global observation-quality map ($\mathrm{F}^\mathrm{global}$). The average runtime is reported in Table~\ref{table-frontier}. $\mathrm{F}^\mathrm{local}$ outperforms $\mathrm{F}^\mathrm{global}$ in both runtime and variance, as the local map contains significantly fewer voxels, reducing the cost of voxel-state queries. Despite requiring additional pose retrieval and visibility checks, $\mathrm{F}^\mathrm{local}$ achieves consistently lower computational costs across all scenarios.

\subsubsection{Incremental Viewpoint Clustering}
To assess the effectiveness of incremental viewpoint clustering, we integrate the proposed module into SSearcher by replacing its original re-clustering strategy, yielding a variant denoted as $\mathrm{SSearcher^{IVC}}$. As summarized in Table~\ref{tab-cmp}, $\mathrm{SSearcher^{IVC}}$ reduces the candidate target generation time (Can. Tm., including viewpoint generation and clustering) by 4.4--32.9$\times$ compared to the original SSearcher. This substantial reduction translates into a higher average velocity and improved target recognition rate, highlighting the benefits of incremental clustering for real-time decision-making.

\newcolumntype{Y}{>{\centering\arraybackslash}X}
\begin{table}[!t]
\centering
\captionsetup{font=small}
\caption{Frontier Detection Time comparisons (ms)}
\label{table-frontier}
\renewcommand\arraystretch{1.1}
\begin{threeparttable}
\begin{tabularx}{\columnwidth}{l *{3}{Y}}
\bottomrule[0.8pt]
\bottomrule[0.8pt]
\textbf{Method} & \textbf{Forest} & \textbf{Garage} & \textbf{Campus} \\
\midrule
$\mathrm{F}^{\mathrm{local}}$ & $4.80\pm1.79$ & $5.86\pm2.74$ & $5.29\pm2.99$ \\
$\mathrm{F}^{\mathrm{global}}$ & $7.29\pm9.28$ & $16.17\pm24.23$ & $9.86\pm13.8$ \\
\bottomrule[0.8pt]
\bottomrule[0.8pt]
\end{tabularx}
\end{threeparttable}
\end{table}

\subsection{Real-World Experiments}  \label{sec-real}
\begin{figure}[h]
\centering
{\includegraphics[width=0.45\textwidth]{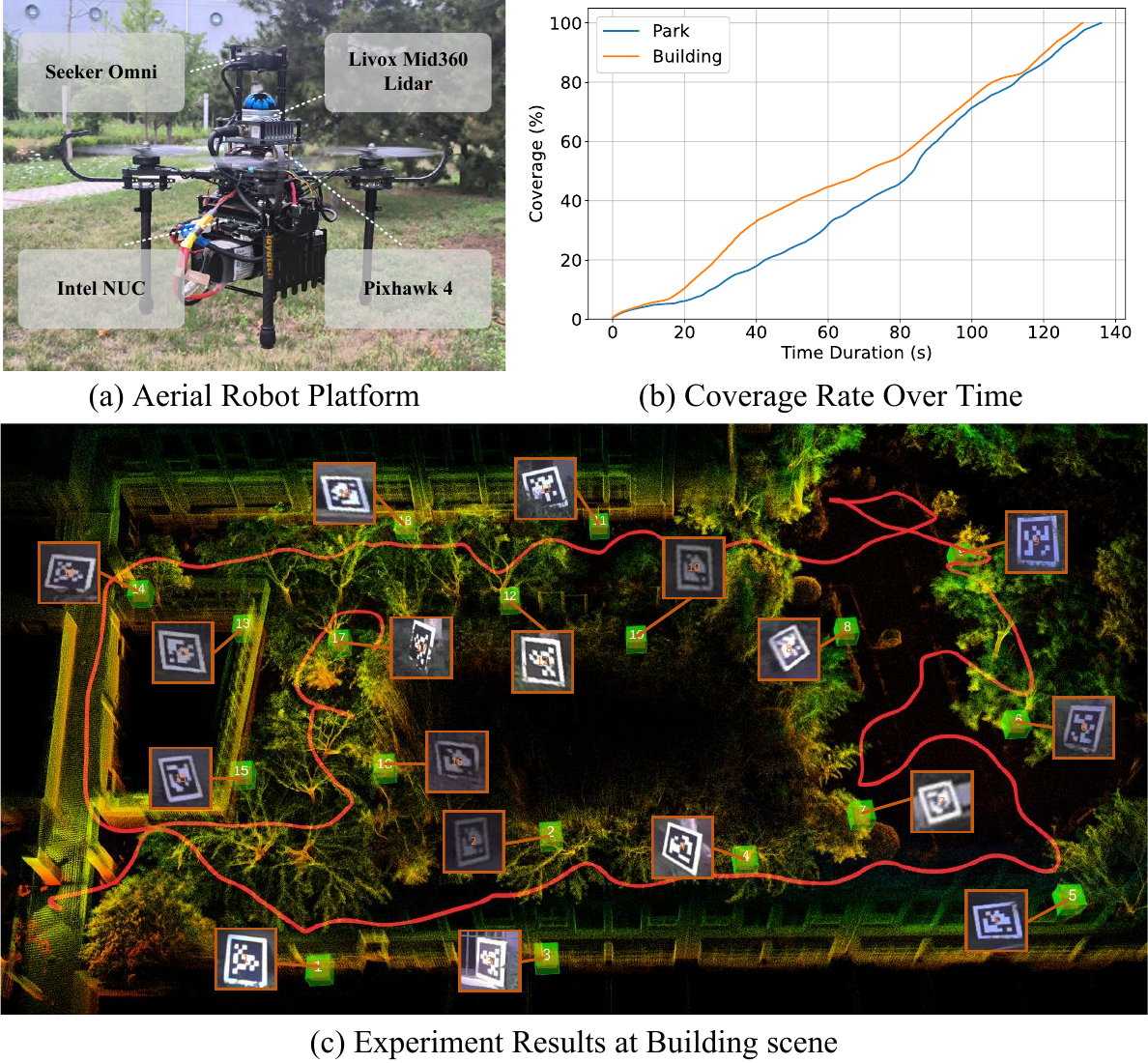}}
\captionsetup{font={small}}
\caption{Experiment results of real-world exploration. Additional visualizations are available on the project page.}
\label{fig_real}
\vspace{-10pt}
\end{figure}

To further assess the proposed algorithm, we conducted real-world experiments with an aerial robot equipped with a MID360 LiDAR and a Seeker Omni camera (Fig.~\ref{fig_real}(a)). FAST-LIO2 \cite{ref-fastlio2} provided onboard state estimation, and a geometric controller \cite{ref-uavctl} handled trajectory tracking, with the maximum velocity limited to $2.5~ \mathrm{m/s}$ for safety. Target search tasks were performed in two environments: a park ($50 \times 35 \times 3~\mathrm{m}^3$) and a building ($60 \times 30 \times 3~\mathrm{m}^3$). In both scenarios, the robot successfully detected all 18 randomly placed targets. The park search took $124.9~ \mathrm{s}$ with a flight distance of $244.2~ \mathrm{m}$ (Fig.~\ref{fig_cover}), while the building search took $124.1~ \mathrm{s}$ and $240.1~ \mathrm{m}$ (Fig.~\ref{fig_real}(c)).

\section{CONCLUSIONS} \label{sec-conclusion}
In this letter, we present SLIDER, an efficient aerial target search framework for large-scale, cluttered environments. By replacing dense global maps with a local sliding map and a sparse global topological map, SLIDER ensures memory- and computation-efficient operations. It integrates real-time observation quality evaluation and incremental viewpoint clustering to achieve adaptive, low-latency planning with long-horizon guidance. Extensive experiments demonstrate that SLIDER outperforms representative baselines in memory usage, planning speed, and search performance. While our lightweight design entails theoretical limitations in extreme occlusion or deadlock scenarios, future work will explore conservative free-space estimation to mitigate observation misjudgments and enhance exploration robustness.
\bibliography{steam} 

\end{document}